\documentclass[conference]{IEEEtran}
\IEEEoverridecommandlockouts
\usepackage{cite}
\usepackage{amsmath,amssymb,amsfonts}
\usepackage{graphicx}
\usepackage{textcomp}
\usepackage{xcolor}
\usepackage[a4paper, total={184mm,239mm}]{geometry}
\usepackage{url}
\usepackage[noend]{algpseudocode}
\usepackage{booktabs,siunitx,array,threeparttable}
\usepackage{xspace}
\usepackage[cjk]{kotex}
\usepackage{kotex-logo}
\usepackage{xcolor}
\usepackage{enumitem}
\usepackage{pifont}
\usepackage[square,sort,comma,numbers,compress]{natbib}
\usepackage{algorithm}
\usepackage{algpseudocode}
\usepackage{multirow}
\usepackage{multicol}
\usepackage{diagbox}
\usepackage{hyperref}
\hypersetup{
    colorlinks=true,
    linkcolor=blue,
    filecolor=magenta,      
    urlcolor=green,
    citecolor=magenta,
    pdfpagemode=FullScreen,
    }
\newcommand{\coin}{M3\xspace}

\newlist{todolist}{itemize}{10}
\setlist[todolist]{label=$\square$}
\usepackage{pifont}

\definecolor{ForestGreen}{RGB}{34,139,34}

\def\BibTeX{{\rm B\kern-.05em{\sc i\kern-.025em b}\kern-.08em
    T\kern-.1667em\lower.7ex\hbox{E}\kern-.125emX}}
\begin{document}


\title{\coin:  Mamba-assisted Multi-Circuit Optimization via MBRL with Effective Scheduling}
\author{\IEEEauthorblockN{Youngmin Oh$^1$, Jinje Park$^1$, Seunggeun Kim$^2$, Taejin Paik$^1$, David Pan$^{2}$, Bosun Hwang$^{1, *}$\thanks{$^*$Corresponding Author: bosun.hwang@samsung.com}
\IEEEauthorblockA{$^1$Samsung Advanced Institute of Technology,\quad $^2$The University of Texas at Austin}
\IEEEauthorblockA{$^1$\{youngmin0.oh, jinje.park, tj.paik, bosun.hwang\}@samsung.com,\quad $^2$\{sgkim@utexas.edu, dpan@ece.utexas.edu\}}
}}

\maketitle

\begin{abstract}
Recent advancements in reinforcement learning (RL) for analog circuit optimization have demonstrated significant potential for improving sample efficiency and generalization across diverse circuit topologies and target specifications. However, there are challenges such as high computational overhead, the need for bespoke models for each circuit. To address them, we propose \coin, a novel Model-based RL (MBRL) method employing the Mamba architecture and effective scheduling. The Mamba architecture, known as a strong alternative to the transformer architecture, enables multi-circuit optimization with distinct parameters and target specifications. The effective scheduling strategy enhances sample efficiency by adjusting crucial MBRL training parameters. To the best of our knowledge, \coin is the first method for multi-circuit optimization by leveraging both the Mamba architecture and a MBRL with effective scheduling. As a result, it significantly improves sample efficiency compared to existing RL methods\footnote{This is a preprint that has not yet completed a peer-review process.}.


\end{abstract}



\section{Introduction}

Analog circuit optimization is one of the most challenging and critical processes in chip design, as it significantly affects the product's reliability and performance. This process is time-consuming due to the vast parameter space and the complex trade-offs between performance metrics. To efficiently automate the process, reinforcement learning (RL) with deep neural networks has recently gained significant attention in analog circuit design~\citep{settaluri2020autockt, wang2020gcn, dac23RoSE, dac23ma},
primarily due to its ability to generalize across varying target specifications~\cite{settaluri2020autockt} and circuit topologies~\cite{wang2020gcn}.

Although RL-based methods have demonstrated remarkable performance in analog circuit optimization, previous studies have typically trained bespoke models for each individual circuit. In other words, to the best of our knowledge, it remains unclear whether it is truly necessary to prepare a separate RL agent for each circuit, as is done by human circuit designers. While some studies have addressed varying target specifications~\cite{settaluri2020autockt}, it is curious to investigate whether a single RL agent can effectively handle multiple circuit topologies and target specifications simultaneously.

\begin{figure}[t!]
\centering
\begin{tabular}{c}    
\includegraphics[width=0.9\columnwidth]{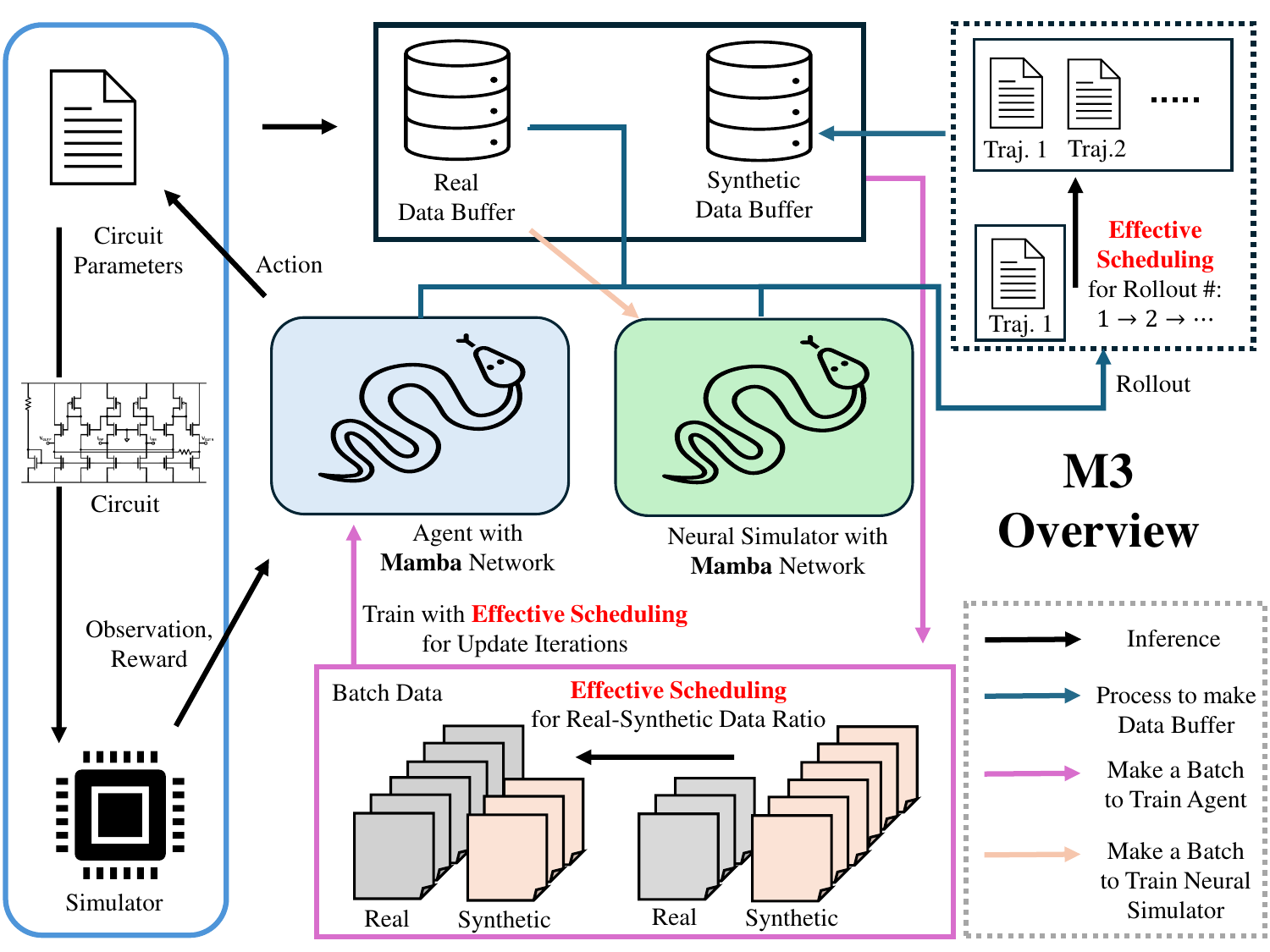}
\end{tabular}
\caption{ \coin overview: The Mamba architecture is utilized for both the agent (i.e., actor and critic networks) and the environment model (i.e., neural simulator). \coin adjusts crucial model-based RL training parameters such as the real-synthetic data ratio, the number of training iterations per environment step, and the number of rollouts. This effective scheduling aims to enhance exploration in the initial stages and maximize exploitation in the later phases.
}
\label{fig:overview}
\end{figure}

Indeed, optimizing multiple circuits using a single neural network (NN) offers significant advantages in terms of efficiency and resource utilization. Even if circuits operate independently and do not interact, concurrent optimization through a unified network enables the parallel processing of various circuits, thereby reducing overall optimization time. This approach allows the network to handle the unique variables and constraints of each circuit within a single framework, leading to more scalable and faster optimization. Furthermore, utilizing a single NN eliminates the need for managing multiple separate models, reducing computational overhead, memory usage, and implementation complexity, which is particularly beneficial in large-scale or complex system-level designs.

Several learning-based multi-circuit optimization approaches exist. One potential candidate involves leveraging graph neural networks to encode a circuit’s graph topology into a fixed-length context vector, which can then be integrated with existing optimization methods. However, this is computationally intensive, especially for large-scale circuits, and lacks a well-established standard to ensure robust performance in parsing circuits into graph representations. As a result, developing a simpler neural network architecture that employs vector-based context representations may offer a more practical alternative, provided it achieves comparable learning performance.

Motivated by the reasons above, we propose a novel framework called \textit{Mamba-assisted Multi-Circuit Optimization via Model-based RL with Effective Scheduling} (\coin). This leverages Mamba~\cite{mamba1, mamba2}, an auto-regressive neural network architecture regarded as a next-generation alternative to the transformer architecture. Compared to the transformer, the Mamba offers high-speed inference with a linear complexity in terms of the data length, while maintaining constant memory usage. Despite these advantages, it is known that the Mamba has comparable performance  to transformer networks. By utilizing this architecture, \coin can handle multi-circuit with distinct parameters and specifications across different dimensions. Furthermore, to reduce the number of interactions with the real simulator, we employ a model-based RL (MBRL) with effective scheduling. Unlike model-free RL, MBRL involves a neural network model to mimic the environment (e.g., simulator) and generate a large volume of synthetic experiences to train a RL  agent.

However, MBRL can sometimes struggle to train the RL agent when the environment is complex, particularly in multi-task scenarios~\cite{yu2020meta} such as multi-circuit optimization. To address this issue, inspired by~\cite{schedule1, schedule3}, \coin gradually adjusts the crucial MBRL training parameters including the real-synthetic data ratio in batch, the number of agent update iteration per environment step, and the number of rollouts to generate  synthetic trajectories. The scheduling is effectively designed as follows: In the initial phase of training, it mainly relies on a large amount of synthetic data, maybe inaccurate yet, to explore circuit parameters getting close to the target specifications (exploration). Later, its tendency changes to predominantly use real data for fine-tuning the policy (exploitation). Fig.~\ref{fig:overview} provides an overview of \coin.

\textbf{Contributions.} We summarize our contributions as follows: \begin{itemize}[leftmargin=0.35cm] 
\item To the best of our knowledge, \coin is the first online learning framework designed for multi-circuit optimization. 
\item \coin employs MBRL with effective scheduling, gradually adjusting crucial  training parameters to address the exploration-exploitation dilemma, achieving high sample efficiency. 
\item \coin is the first method to leverage the Mamba architecture for effectively handling multi-circuit with distinct circuit parameters and target specifications. 
\item We validate \coin by benchmarking it against other RL baselines across various analog circuits to show significant improvements in sample efficiency. \end{itemize}


\section{Related Works}

\textbf{Mamba.} Recently, a new architecture, known as Mamba, has emerged as one of the candidates replacing the transformer architecture.
The Mamba yields powerful results due to its strong long-term memory retention and efficiency. For instance, 
Lieber et al. introduced Jamba~\cite{jamba}, which covers a context length that is 7 times more than that of Llama2-70B~\cite{llama2} with a comparable performance. Besides, due to its linear scalability, high potential, and versatility, the Mamba architecture is being studied across various field such as natural language processing~\cite{jamba}, 
time-series forecast~\cite{mambaformer}, 
vision~\cite{visionmamba}.
 For more detailed information, refer to \cite{surveymamba}. As far as we know, however, Mamba has yet to be applied in analog circuit optimization.

\textbf{Reinforcement Learning.} Reinforcement learning (RL) methods are categorized into model-free RL (MFRL) and model-based RL (MBRL) approaches. Recent well-known MFRL methods, such as PPO~\cite{ppo}, SAC~\cite{sac}, learn an optimal policy through direct interaction with the environment; therefore, they do not require a surrogate model that describes the environment. However, because these methods require a large number of direct interactions, they are not suitable in cases where direct interaction is time-consuming or costly. On the other hand, MBRL methods such as MBPO~\cite{mbpo}, BMPO~\cite{bmpo}, and M2AC \cite{m2ac}, utilize the surrogate model that mimics the environment. The surrogate model is used to generate synthetic data indistinguishable from the real data, enabling a RL agent to learn a good policy with a large amount of data. Therefore, the overall number of direct interactions throughout the learning process is significantly reduced if the surrogate model is successfully trained. Furthermore, \cite{schedule1, schedule3} studied that effective scheduling for crucial MBRL training parameters, such as the real-synthetic data ratio in a batch and the number of rollouts, largely affects the performance of MBRL policies. In particular, \cite{schedule1} reported that gradually increasing the proportion of real data results in improved performance. However, to the extent of our knowledge, no study has yet explored MBRL with effective scheduling for addressing the optimization of multi-circuit belonging to the context of multi-task RL. Inspired by prior work, we also adopt this methodology.

\textbf{Transistor Sizing.} 
Given the complexity of modern circuit designs, genetic algorithms and Bayesian optimization have been developed to efficiently explore the design space 
\cite{kongpvtsizing,liu2021parasitic}.
Large language models also contribute by automating code generation and EDA script writing, streamlining the design process \cite{chang2024data}. In particular, 
RL has emerged as a promising approach for learning optimal design parameters through interactions with simulation environments.
However, MFRL methods are significantly limited by the high computational costs required by simulations. 
MBRL, which leverages surrogate models for the simulator, presents a more efficient alternative by reducing the simulator's computational overhead while optimizing key circuit metrics like power, area, and delay \cite{cronus, ahmadzadeh2024using}. For instance, CRoNuS \cite{cronus}, a MBRL approach that replaces costly simulators with an ensemble of neural networks, demonstrated over $5\times$ higher sample efficiency in optimizing single analog circuit designs. Besides, \cite{poddar2024insight} used the transformer to train a neural simulator handling multi-circuit prediction. However, to the best of our knowledge, no prior work has explored using Mamba and MBRL with effective scheduling for multi-circuit optimization.


\vspace{-0.1in}
\section{Method}
\subsection{Reinforcement Learning Formulation}

We consider a set of $C$ circuits, indexed by $C=\{1, \dots, C\}$. Let $N_c$ represent the number of circuit parameters (e.g., transistor width, multiplier, resistor) and $K_c$ denote the number of performance metrics (e.g., gain, bandwidth, phase margin) for each circuit $c \in C$. Following \cite{cronus, settaluri2020autockt}, we define the observation, action, and reward spaces, which are fundamental components for formulating a Markov decision process (MDP) in reinforcement learning (RL).

\subsubsection{Observation}
The observation for a circuit $c$ is defined as $\mathbf{o}_c = \mathbf{o}_{c,e} \oplus \mathbf{o}_{c,v}$, with  a circuit embedding $\mathbf{o}_{c,e}$ and
\begin{align}
    \mathbf{o}_{c,v} &= \left(\bigoplus_{i=1}^{N_c} p_{c,i}\right) \oplus \left(\bigoplus_{i=1}^{K_c} \text{d}(m_{c,i}, g_{c,i}) \oplus \text{d}(n_{c,i}, g_{c,i})\right),\nonumber
\end{align}
with the function $\text{d}(x, y) = \frac{x - y}{x + y}$. Here, $p_c = \{p_{c,1}, \dots, p_{c,N_c}\}$ represents the current design parameters, $\{m_{c,1}, \dots, m_{c,K_c}\}$ the performance metrics, $\{n_{c,1}, \dots, n_{c,K_c}\}$ the target specifications, and $\{g_{c,1},\dots,g_{c,K_c}\}$ the normalizing factors. In other words, $p_{c,i}$ is the $i$-th circuit parameter (e.g., multiplier), and $m_{c,i}$ and $n_{c,i}$ are the $i$-th corresponding current and target specifications (e.g., gain). The operator $\oplus$ denotes the concatenation.

\subsubsection{Action} Next, we define the action for circuit $c$ as the change in its parameters, represented by $\Delta p_c$. The updated parameters are then given by $p_c \leftarrow p_c + \Delta p_c$. The action is scaled to lie within the range $[-1, 1]^{N_c}$, enabling the learning agent to efficiently explore the parameter space. This scaling ensures that the full range of circuit parameters can be covered within each episode when mapped back to their original values.

\subsubsection{Reward} The reward function $r$ for a circuit $c \in C$ is defined as follows as \cite{settaluri2020autockt, cronus}:

\begin{align}
r\left(\bigoplus_{i=1}^{K} m_{c,i} \oplus \bigoplus_{i=1}^{K} n_{c,i}\right) &= 
\begin{cases} 
\texttt{FoM} & \text{if } \texttt{FoM} < -0.02, \\
10 & \text{if } \texttt{FoM} \geq -0.02,
\end{cases}\\
\texttt{FoM} &= \sum_{i=1}^{K} \min\left\{\text{d}(m_{c,i}, n_{c,i}), 0\right\}.
\end{align} 

The different dimensions of the action and observation spacesacross circuits $c$  require padding to standardize the input and output lengths. Handling these packed inputs, we were able to utilize the autoregressive models for multi-circuit optimization. 

\subsection{Actor, Critic, and Model Networks}

\textbf{Motivation of Mamba:}
The transformer architecture, widely used across various domains due to its powerful focus and generalization capabilities through the attention module, faces challenges when handling long input sequences during inference. This limitation stems from the fact that the computational cost of the attention mechanism grows quadratically with input length. In contrast, the Mamba architecture has demonstrated an ability to efficiently capture sequential dependencies within input data while maintaining linear scalability relative to input length. Along with this scalability, Mamba leverages optimized parallel processing techniques, allowing it to outperform transformers in certain domains.

\begin{figure}[!t]
\centering
\begin{tabular}{c}    
\includegraphics[width=0.9\columnwidth]{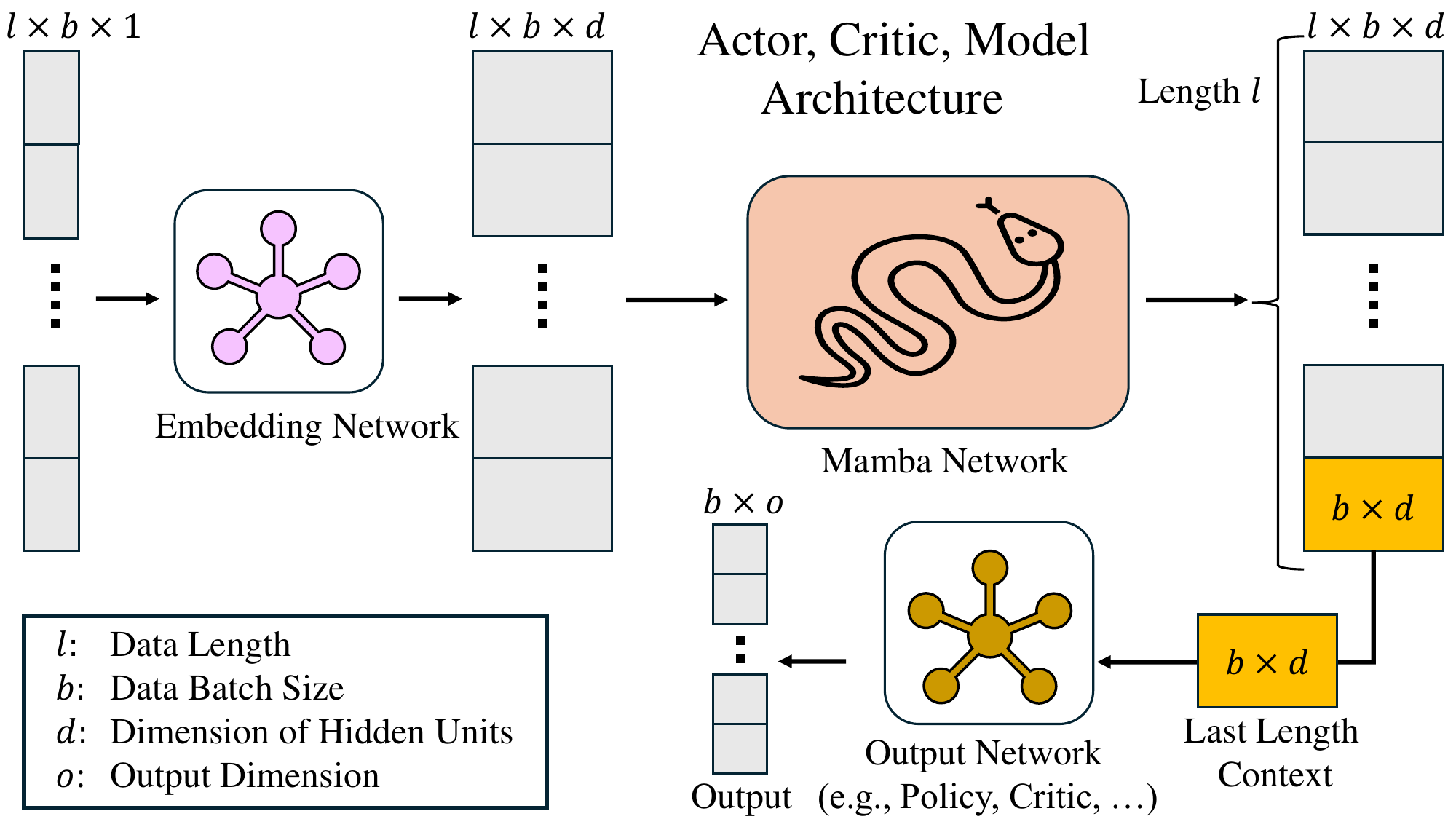}
\end{tabular}
\caption{Architecture of Neural Networks. $o$ is output size, i.e., the dimension of outputs, so that it depends on the objective of neural networks; for instance, $o=1$ for critic ($Q$-network) since $Q$-values are real-valued.
}
\label{fig:arch}
\end{figure}

\textbf{Implementation:}
Let $x$ be a batch of packed observations, actions, or tuples of observation and action with batch size $b$. Then $x\in\mathbb{R}^{b\times l}$, e.g., $l$ could be the dimensions of packed observation, or action, respectively. Then we can regard $l$ as a full length, i.e., x is data $\mathbb{R}^{b\times l\times 1}$ for length $l$ with the dimension $1$. For each network $X \in \{\text{actor, critic, environment model}\}$, an embedding network $E_X: \mathbb{R} \rightarrow \mathbb{R}^{d}$ and the Mamba network $M_X$ are constructed such that $E_X(x^\intercal) \in \mathbb{R}^{l \times b \times d}$ and $y = \{y_1, \cdots, y_l\} = M_X(E_X(x^\intercal)) \in \mathbb{R}^{l \times b \times d}$. We then use the final context vector $y_l$ as input to compute $L(\sigma(y_l)): \mathbb{R}^d \rightarrow \mathbb{R}^{d'}$, where $\sigma$ is an activation function and $L$ is a linear layer. The output dimension $d'$ depends on the role of the neural network. Specifically, for observation and action dimensions $d_o$ and $d_a$, $d'$ equals $d_a$, 1, and $d_o + 1$ for the actor $\pi_\theta$, critic $Q_\theta$, and model networks $\{\mathcal{N}_\theta\}_{i=1}^{N_{\text{total}}}$, respectively, where $\theta$ are the learnable parameters. Fig.~\ref{fig:arch}
 illustrates the architecture of the actor, critic, and model networks.
 
\begin{algorithm}[t!]
\caption{PseudoCode for \coin}
\begin{algorithmic}[1]
\Require Set environment $\mathcal{E}$, $N_{total}$, $N_{elite}$, $b$, $\alpha_{I}, \alpha_{F},$ $N_I$, $T_{model}$, $T_{a,I}, T_{a,F}, T_{\max}$, $R_I$, $R_F$, $T_{ro}$
\State Initialize actor $\pi_{\theta}$, critic $Q_{\theta}$, and the ensemble of model networks $\{\mathcal{N}_\theta\}_{i=1}^{N_{total}}$ for learnable parameters $\theta$
\State Collect data into a buffer $\mathcal{B}_{real}$ by $N_I$ random actions
\For{$t = 1$ to $T_{\max}$}
    \If{$t \equiv 0 \pmod{T_{model}}$}
        \State Train $\{\mathcal{N}_\theta\}_{i=1}^{N_{total}}$ by using data in $\mathcal{B}_{real}$.
        \State Define a set of $N_{elite}$ elites from $\{\mathcal{N}_\theta\}_{i=1}^{N_{total}}$
        \State Rollout by the elites and $\pi_{\theta}$ 
    \EndIf
    \If{$t \equiv 0 \pmod{T_{ro}}$}
        \State Update the rollout number $R(t)$ by \eqref{eq:ron}
        \State Construct synthetic data buffer $\mathcal{B}_{sync}$ based on $R(t)$
    \EndIf
    \State Reset $\mathcal{E}$ if necessary to change circuit topology, technology node, and target specifications 
    \State Interact with $\pi_\theta$ and $\mathcal{E}$ to get data, e.g., reward
    \State Push data into $\mathcal{B}_{real}$
    \For{$i=1$ to $T_a(t)$ using \eqref{eq:Ta}}
        \State Make a batch with the ratio $\alpha(t)$ in \eqref{eq:alpha}
        \State Train $\pi_{\theta}$ and $Q_{\theta}$ by the batch
    \EndFor
\EndFor
\end{algorithmic}\label{alg:coin}
\end{algorithm}

\subsection{Training}

\subsubsection{Initial Phase}
We assume the environment satisfies the following conditions: First, whenever the environment resets, a new topology, technology, and target specifications are selected. Second, a reset occurs either when the reward becomes positive or when the episode's time step exceeds the maximum episode length, $T_{ep} > 0$. The actor and critic networks are initialized with random weights, and an ensemble of environment models is created, consisting of a total size $N_{total}$ with an elite subset of size $N_{elite}$. Using randomly selected actions, we collect $N_I$ data, consisting of the current and next observations, actions, rewards, and other relevant information, into a real data buffer through interactions with the real simulator

\subsubsection{Training Phase}
The training process is divided into two parts: model training and agent training.

\textbf{Model Training:} At model learning frequency $T_{model}$, \coin divides the data in the environment buffer into training and validation sets based on a validation data ratio. The ensemble of model networks is trained to predict rewards and next observations, using action and current observation tuples as inputs. The training of the ensemble terminates when no significant improvement is observed in validation performance. Once trained, \coin chooses a set of elites from the ensemble. At rollout frequency $T_{ro}$, a model buffer is newly constructed to collect synthetic data using the actor and the elites, with the initial observations coming from the environment buffer, based on the rollout number $R(t)$ for  hyperparameter $scale>0$:
 
\begin{align}
    R(t) &= \max\left(\min\left(\widetilde{R}(t), R_M\right), R_m\right), \label{eq:ron}    
\end{align} with $\widetilde{R}(t)=R_{I} + t\left(\frac{R_{F} - R_{I}}{scale}\right)$ for the environment step $t$ where $R_I$ and $R_F$ are the initial and last rollout numbers, and $R_m=\min\left(R_I, R_F\right)$, $R_M=\max\left(R_I, R_F\right)$.

\textbf{Agent Training:} 
\coin adopt SAC~\cite{sac}, one of the model-free reinforcement learning (MFRL) algorithms, as the core of policy learning. At each time step $t$, the actor and critic networks are updated with the number of $T_a(t)$. During each iteration, the real-synthetic dat ratio $\alpha(t)$ is used to sample $\alpha(t) b$ real data and $(1 - \alpha(t))b$ synthetic data, respectively, in a batch for a predefined batch size $b$. \coin gradually adjust the values of $T_{a}(t)$ and $\alpha(t)$ by the following relations: for environment step $t$ with hyperparameter $scale>0$,

\begin{align}
    \alpha(t) &= \max\left(\min\left(\widetilde{\alpha}(t), \alpha_{M}\right),\alpha_{m}\right), \label{eq:alpha}\\
    T_a(t) &= \max\left(\min \left(\widetilde{T}_a(t), T_{a,M}\right), T_{a,m}\right), \label{eq:Ta}
\end{align} with 
        $\widetilde{\alpha}(t) = \alpha_{I} + t\left(\frac{\alpha_{F} - \alpha_{I}}{scale}\right)$ and 
    $\widetilde{T}_a(t) = T_{a, I} + t\left(\frac{T_{a, F} - T_{a, I}}{scale}\right)$
where $\alpha_{I}$ and $\alpha_{F}$ are the initial and final values of $\alpha(t)$, and $T_{a, I}$ and $T_{a, F}$ are the initial and final values for $T_a(t)$, respectively. In addition, we set $\alpha_m=\min(\alpha_I, \alpha_F)$, $\alpha_M=\max(\alpha_I, \alpha_F)$ and $T_{a,m}=\min(T_{a,I}, T_{a,F})$, $T_{a,M}=\max(T_{a,I}, T_{a,F})$, respectively.

To conclude, in \coin, both the rollout number \( R(t) \), the agent update iteration number \( T_a(t) \), and the real-to-synthetic data ratio  \( \alpha(t) \) in a batch increase as the number of environment steps progresses. The motivation behind this trend is as follows: Initially, the agent is trained with a substantial amount of synthetic data, which may be inaccurate but is sufficient to guide the agent in efficient exploration. As the environment steps accumulate, more real data are collected, allowing the neural simulator to be well-trained. Thus, to fine-tune the agent for a high-performance policy, higher rollouts and an increased ratio of real data in the training batch will be utilized, along with a greater number of update iterations for exploitation.

The pseudocode for \coin is provided in Algorithm~\ref{alg:coin}.

\section{Experimental Results}
In this section, we answer the following questions to validate \coin's effectiveness:
\begin{enumerate}[label=\textbf{Q\arabic*:}, leftmargin=*]
    \item Is Mamba more effective for multi-circuit optimization compared to fully connected layers and other autoregressive methods, such as transformer?
    \item Is MBRL with effective scheduling is really valid compared to standrd MBRL?
    \item Does \coin outperform other popular RL methods, such as SAC~\cite{sac}, PPO~\cite{ppo}?
\end{enumerate} 

\begin{table}[h!]
\begin{center}
\caption{Main Hyperparameters}
\label{tab:hyperparam}
\vspace{-0.1in}
\resizebox{0.9\columnwidth}{!}{
\begin{tabular}{l|c}
\toprule
Parameter & Value \\
\midrule
Dimension of Model and State in Mamba & 64, 16 \\
Dimension and Expand of ConvNet in Mamba & 4, 2\\
Number of Ensemble Networks (Elites) & 7 (5)\\
$T_{ro}$&30\\
$T_{model}$ & 300 simulator runs\\
Learning Rate & 3e-4\\
$R_I, R_I$ & 1, 7
\\
 $T_{a,I}, T_{a,F}$ & 15, 20\\
 $\alpha_{I}, \alpha_{F}$ & $0.05, 0.95$\\
 $scale$ in \eqref{eq:alpha}&15,000\\
 Patience of Model Training  & 5\\
 Validation Data Ratio of Model Training & 0.2\\
 Gradient Clipping Value & 1.0\\
 Optimizer & Adam Optimizer~\cite{adam}
 
\\
\bottomrule
\end{tabular}}
\end{center}
\vspace{-0.2in}
\end{table}

\begin{table*}[t!]
\centering
\caption{Circuit Target Specification and Parameter Ranges with 45nm technology.}
\label{tab:my-table}
{
\resizebox{0.9\textwidth}{!}{%
\begin{tabular}{|c|cccclc|cc|c|}
\hline
\multirow{2}{*}{} & \multicolumn{6}{c|}{Target Specifications}                                                                                                                                             & \multicolumn{2}{c|}{Parameters}                            & \multirow{2}{*}{Ob. Dim} \\ \cline{2-9}
                  & \multicolumn{1}{c|}{Gain}             & \multicolumn{1}{c|}{BW (Hz)}           & \multicolumn{1}{c|}{PM ($^\circ$)} & \multicolumn{2}{c|}{$I_{\textrm{bias}}$ (A)} & Volrage Swing (V) & \multicolumn{1}{c|}{Multiplier}             & $C$ (F)            &                          \\ \hline
2SOA              & \multicolumn{1}{c|}{{[}2e2, 4e2{]}}   & \multicolumn{1}{c|}{{[}1e6, 2.5e7{]}}  & \multicolumn{1}{c|}{60}            & \multicolumn{2}{c|}{{[}1e-4 ,1e-2{]}}        & 0.5               & \multicolumn{1}{c|}{{[}1, 100{]}$^6$} & {[}1e-13,1e-11{]}  & 19                       \\ \hline
R2SOA             & \multicolumn{1}{c|}{{[}2e2, 4e2{]}}   & \multicolumn{1}{c|}{{[}1e6, 2.5e7{]}}  & \multicolumn{1}{c|}{60}            & \multicolumn{2}{c|}{{[}1e-4, 1e-2{]}}        & None              & \multicolumn{1}{c|}{{[}1, 100{]}$^6$} & {[}1e-13, 1e-11{]} & 15                       \\ \hline
2STIA             & \multicolumn{1}{c|}{{[}2.5e2, 5e2{]}} & \multicolumn{1}{c|}{{[}4.5e9, 1e10{]}} & \multicolumn{1}{c|}{60}            & \multicolumn{2}{c|}{{[}4e-2, 2e-1{]}}        & None              & \multicolumn{1}{c|}{{[}1, 100{]}$^6$} & None               & 14                       \\ \hline
                  & \multicolumn{3}{c|}{Delay (s)}                                                                                      & \multicolumn{3}{c|}{Power (W)}                                   & \multicolumn{2}{c|}{Transistor Width ($\mu$)}                     &                          \\ \hline
Comp              & \multicolumn{3}{c|}{{[}4.5e-12, 9e-12{]}}                                                                           & \multicolumn{3}{c|}{{[}2e-10, 2.5e-10{]}}                        & \multicolumn{2}{c|}{{[}0.1, 100{]}$^6$}                    & 10                       \\ \hline
\end{tabular}
}}\vspace{-0.2in}
\end{table*}

\begin{figure*}[!t]
    \centering
    \resizebox{0.8\textwidth}{!}{
    \begin{tabular}{cccc}
    \includegraphics[width=0.5\columnwidth]{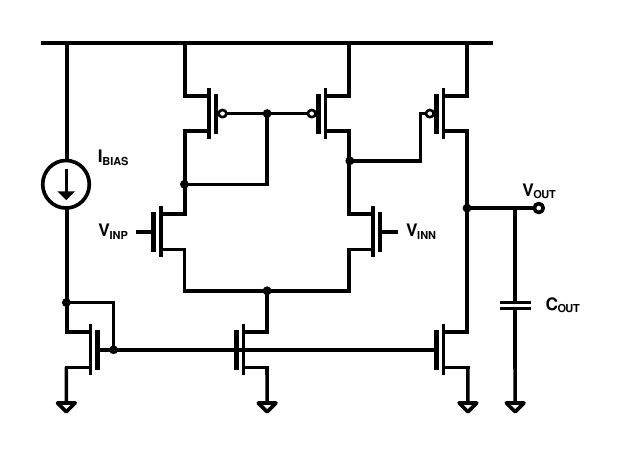}
    &\includegraphics[width=0.5\columnwidth]{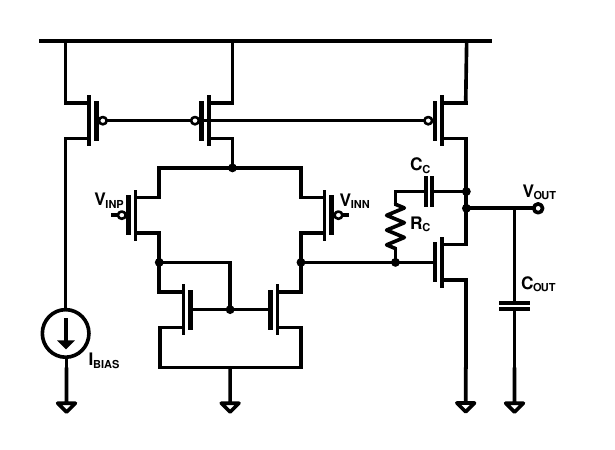}
    &\includegraphics[width=0.5\columnwidth]{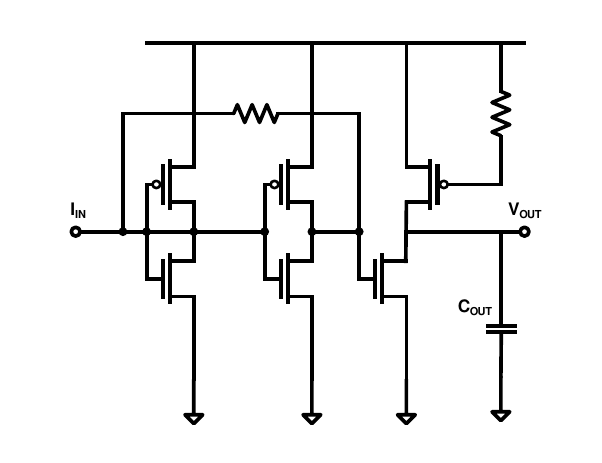}
    &\includegraphics[width=0.5\columnwidth]{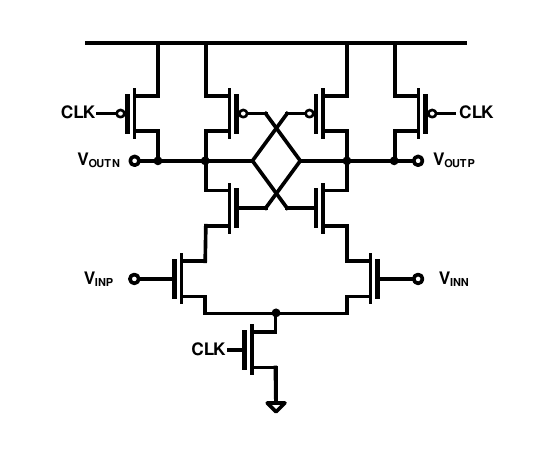}
    \\
    (a)  & (b) & (c) &(d)
    \end{tabular}
    }
    \caption{Benchmark Circuits.  (a) Two-stage operational amplifier (2SOA), (b) Two-stage operational amplifier exchanging p-mos and n-mos transistors each other (R2SOA), (c) Two-stage transimpedence amplifier (2STIA), (d) Comparator (Comp)}    
    \label{fig:circuits}
\end{figure*}

\begin{figure*}[!t]
    \centering
    \resizebox{0.95\textwidth}{!}{
    \begin{tabular}{c}
    \includegraphics[width=\columnwidth]{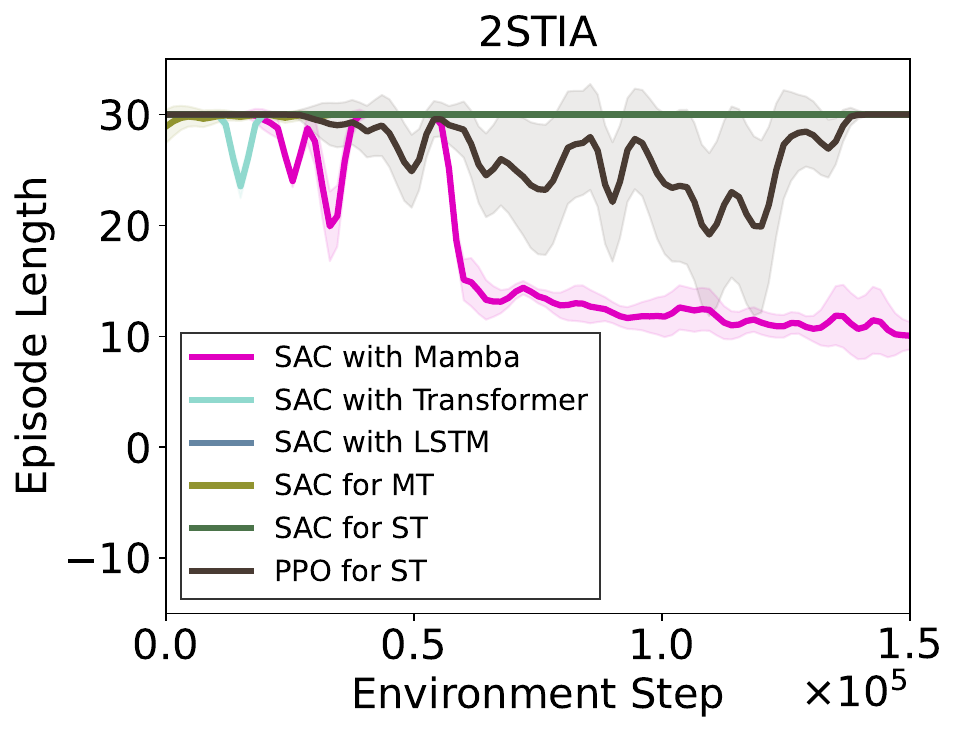}
    \includegraphics[width=\columnwidth]{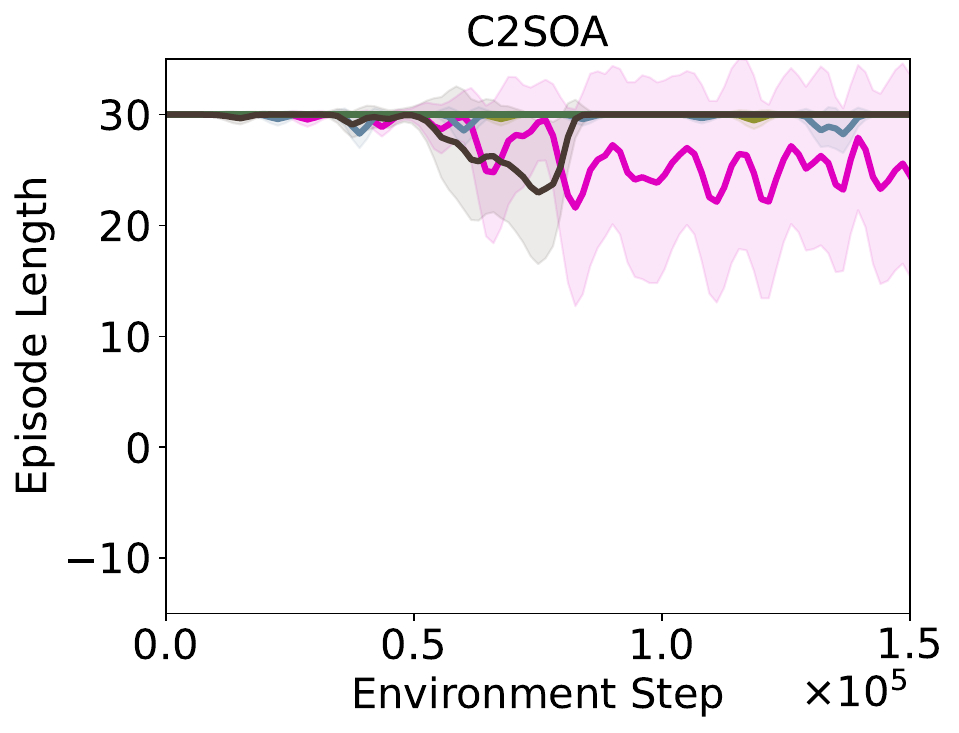}
    \includegraphics[width=\columnwidth]{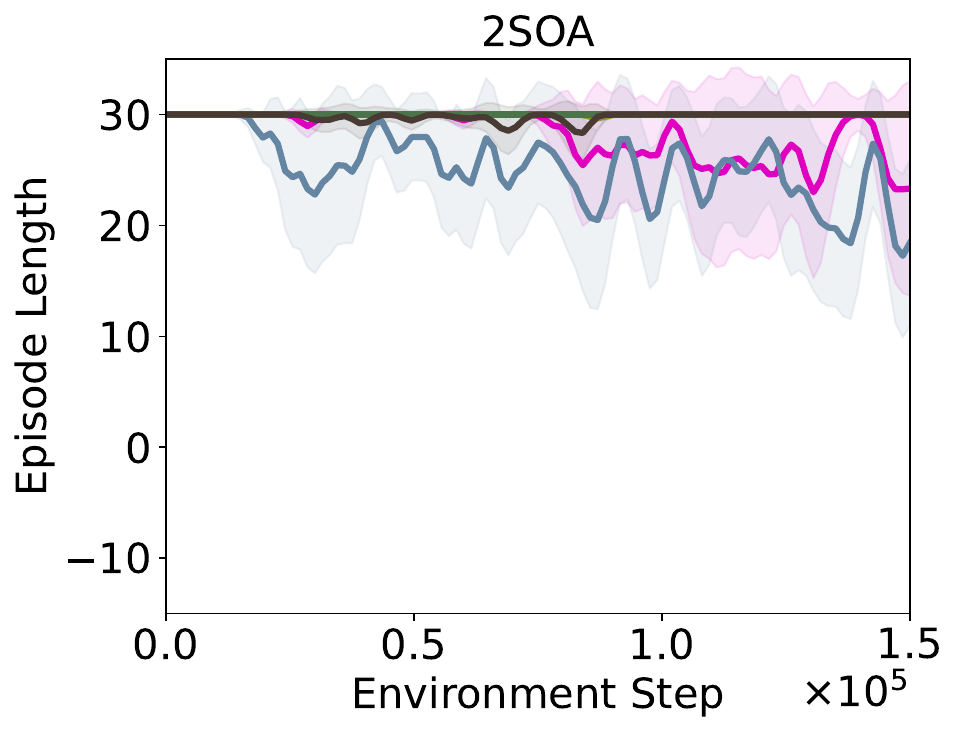}
    \includegraphics[width=\columnwidth]{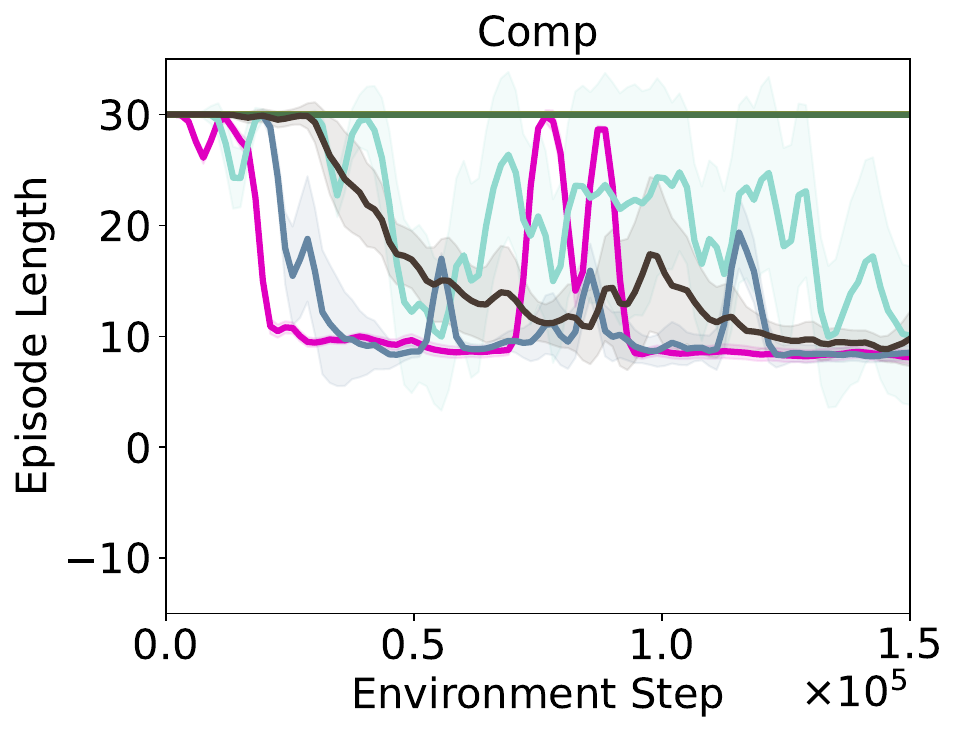}
    \\
     \includegraphics[width=\columnwidth]{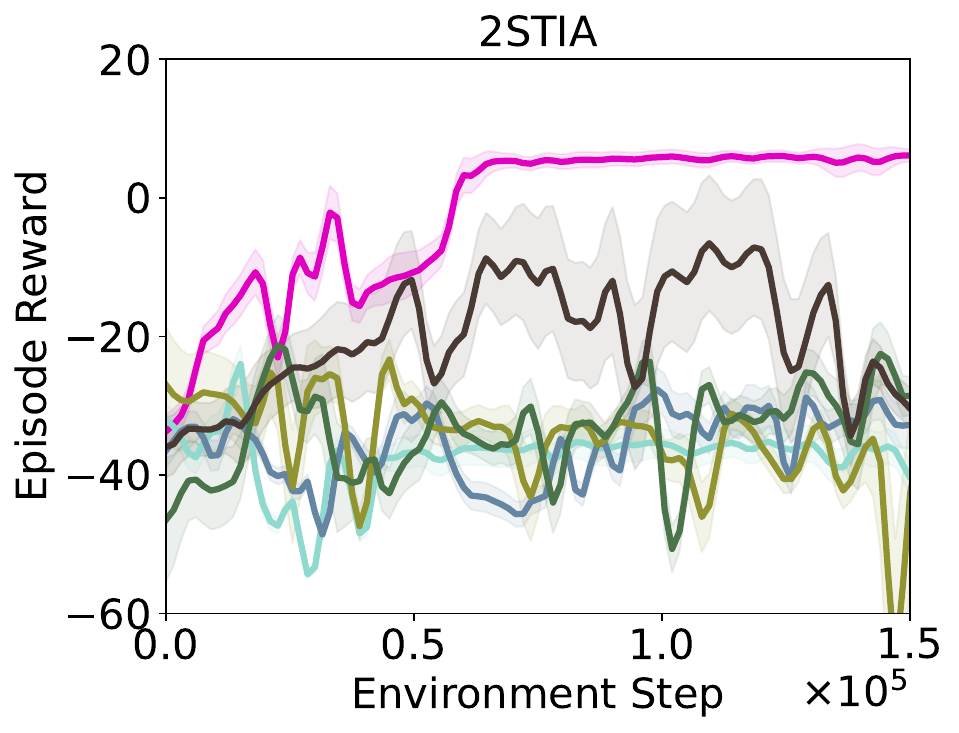}
    \includegraphics[width=\columnwidth]{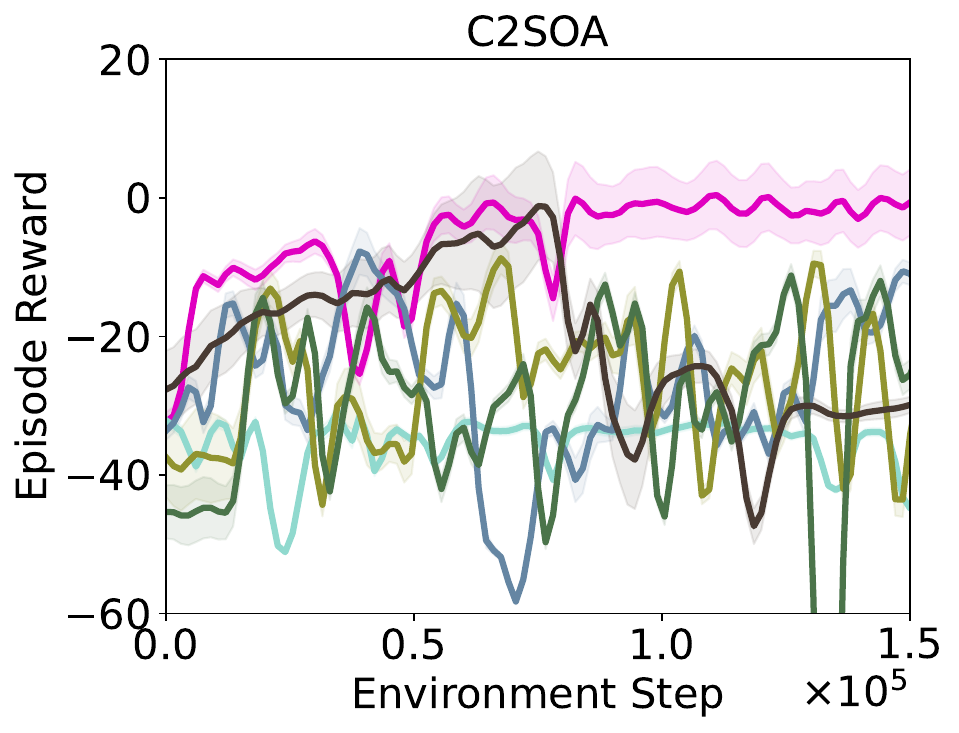}
    \includegraphics[width=\columnwidth]{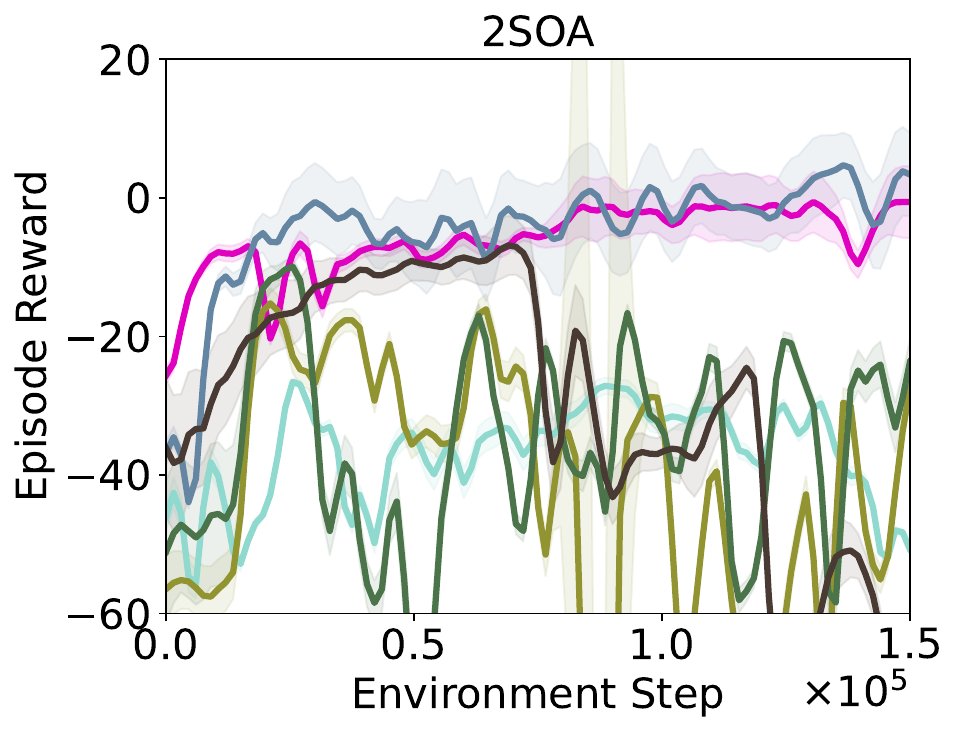}
    \includegraphics[width=\columnwidth]{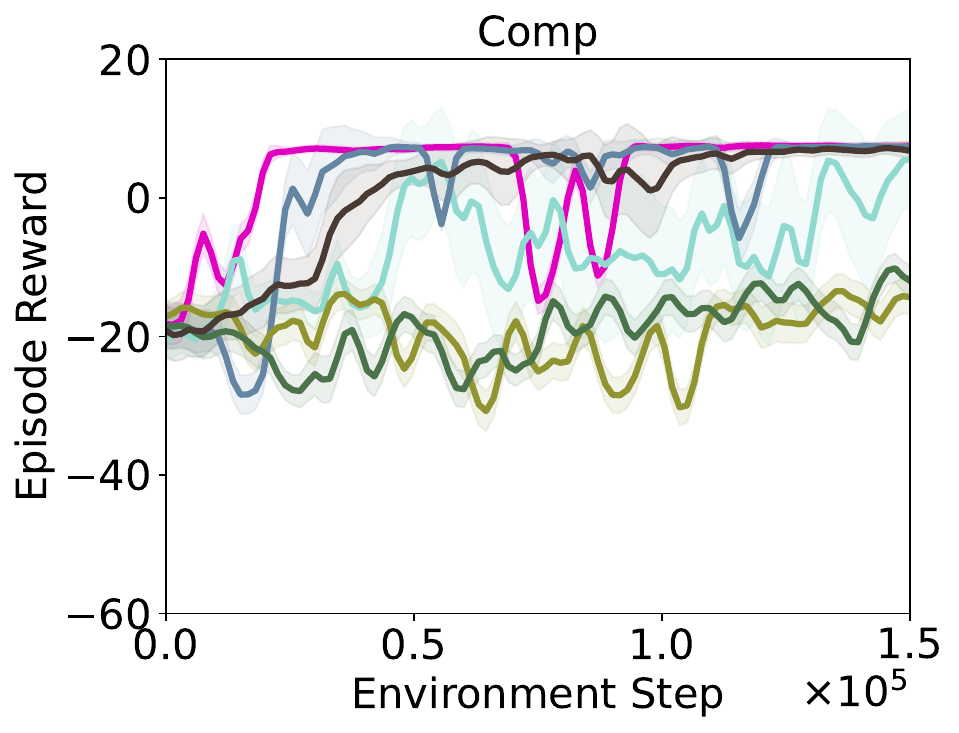}
    \end{tabular}}
    \caption{Learning curves of episode lengths and episode rewards. Both MT and ST denote multi- and single-task learning, respectively. The solid line
and shaded regions represent the mean and standard deviation, respectively, across ten runs  with random seeds.}
    \label{fig:exp1}

\end{figure*}
\begin{figure*}[!t]
    \centering
    \resizebox{0.95\textwidth}{!}{
    \begin{tabular}{c}
    \includegraphics[width=\columnwidth]{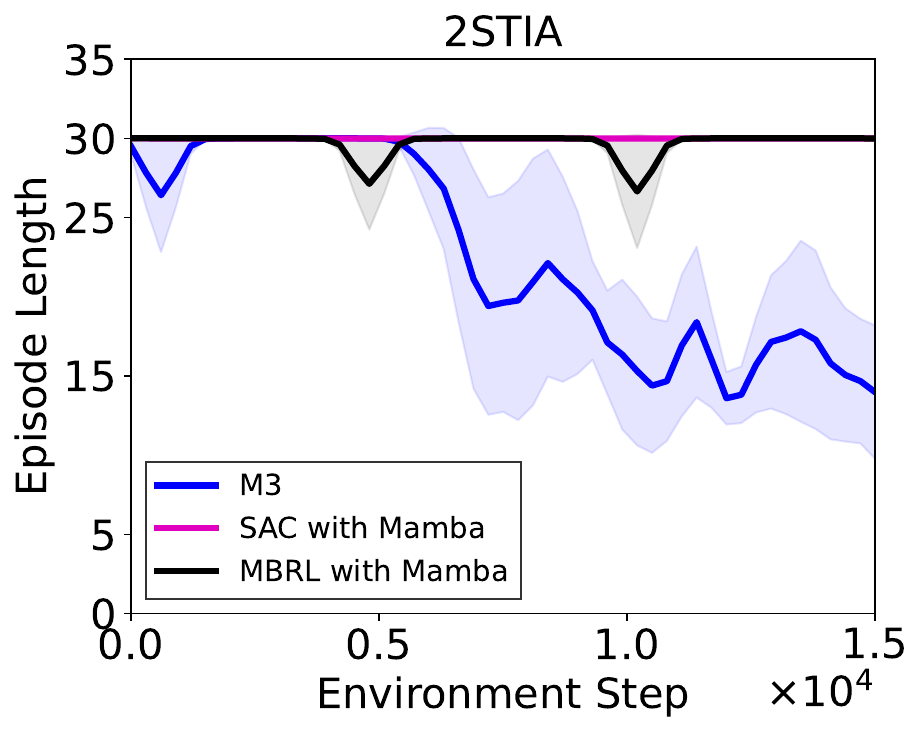}
    \includegraphics[width=\columnwidth]{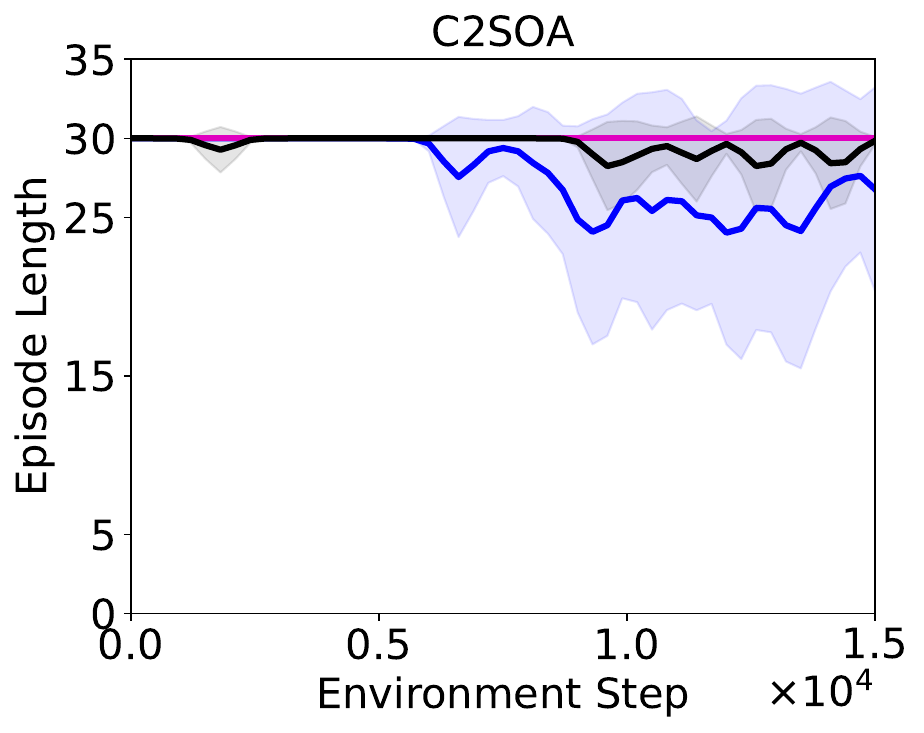}
    \includegraphics[width=\columnwidth]{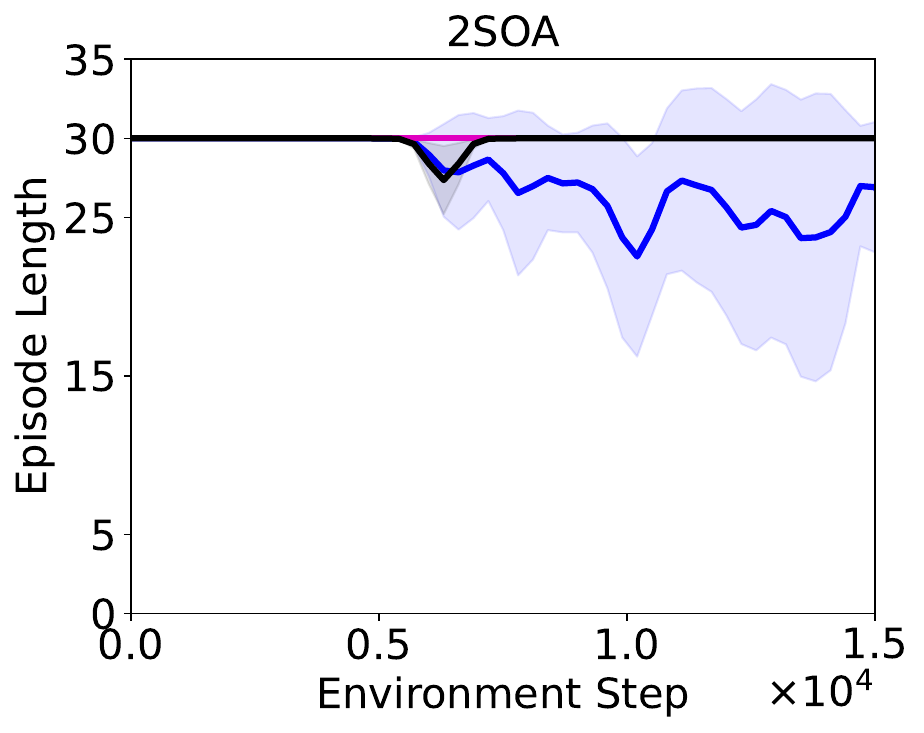}
    \includegraphics[width=\columnwidth]{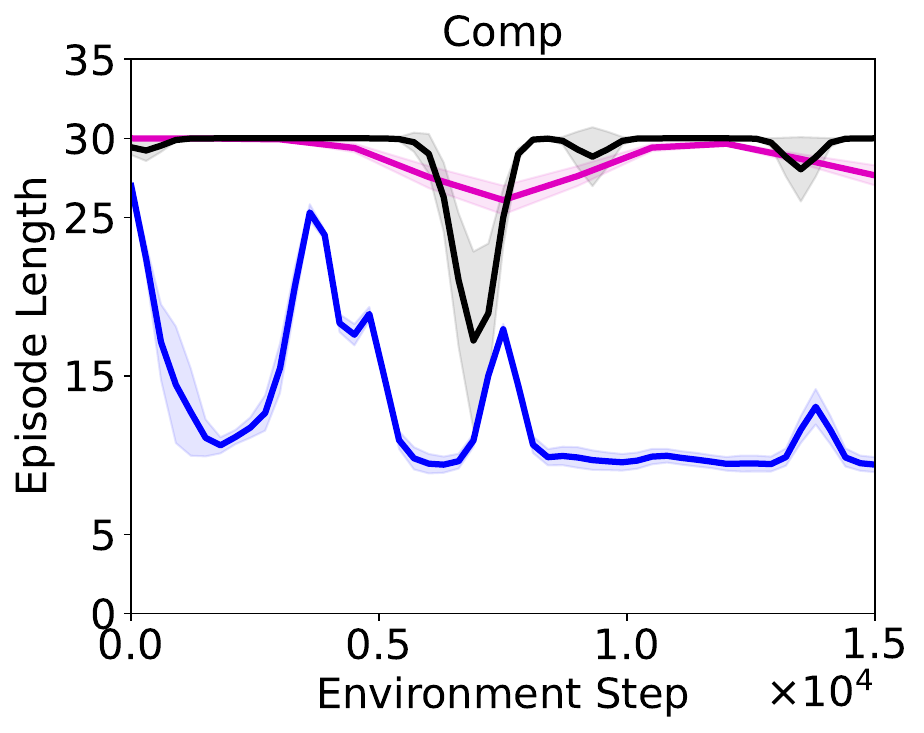}
    \\
     \includegraphics[width=\columnwidth]{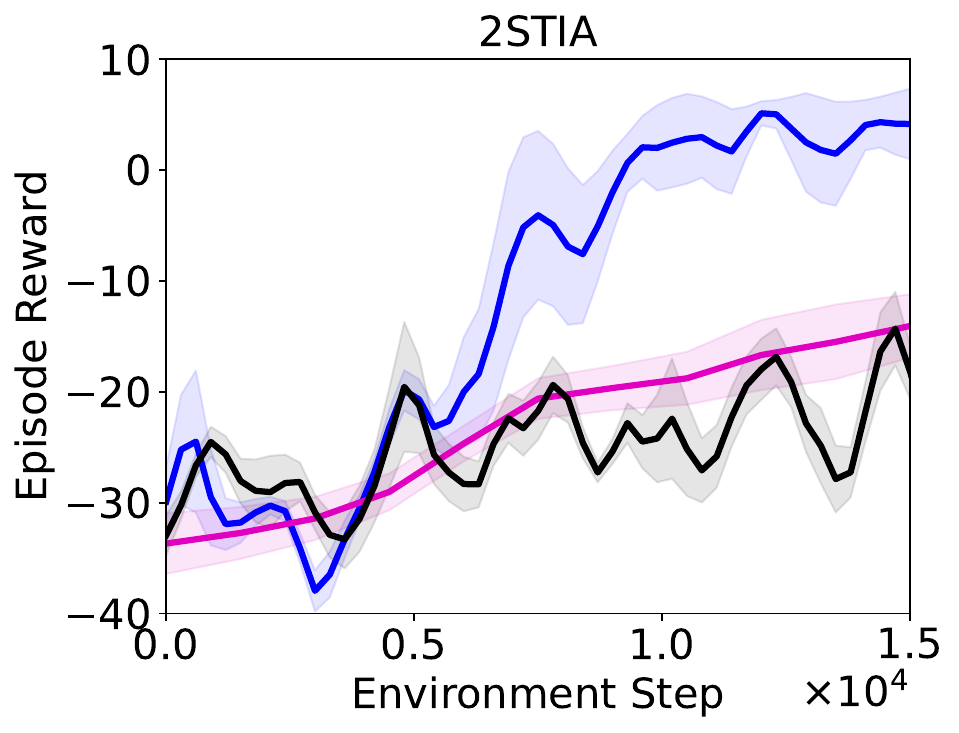}
    \includegraphics[width=\columnwidth]{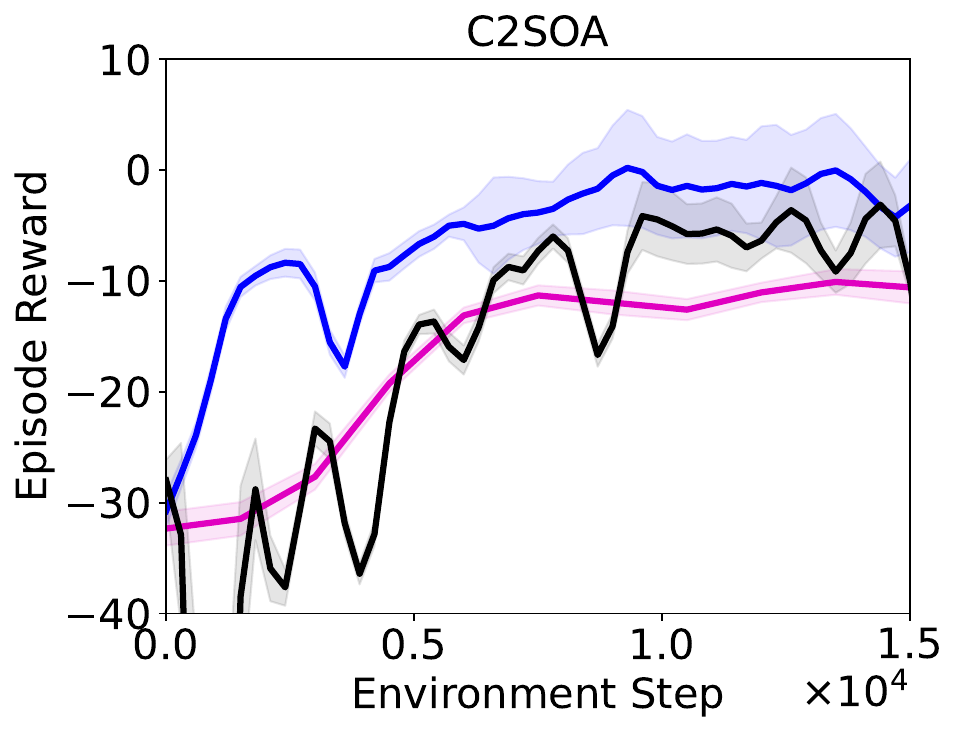}
    \includegraphics[width=\columnwidth]{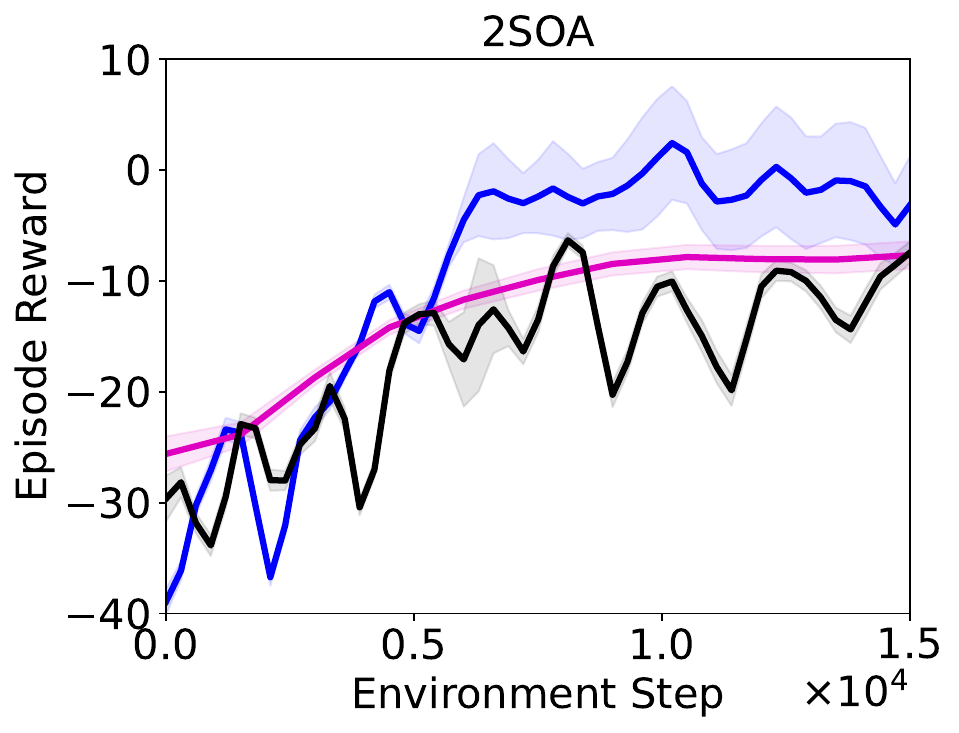}
    \includegraphics[width=\columnwidth]{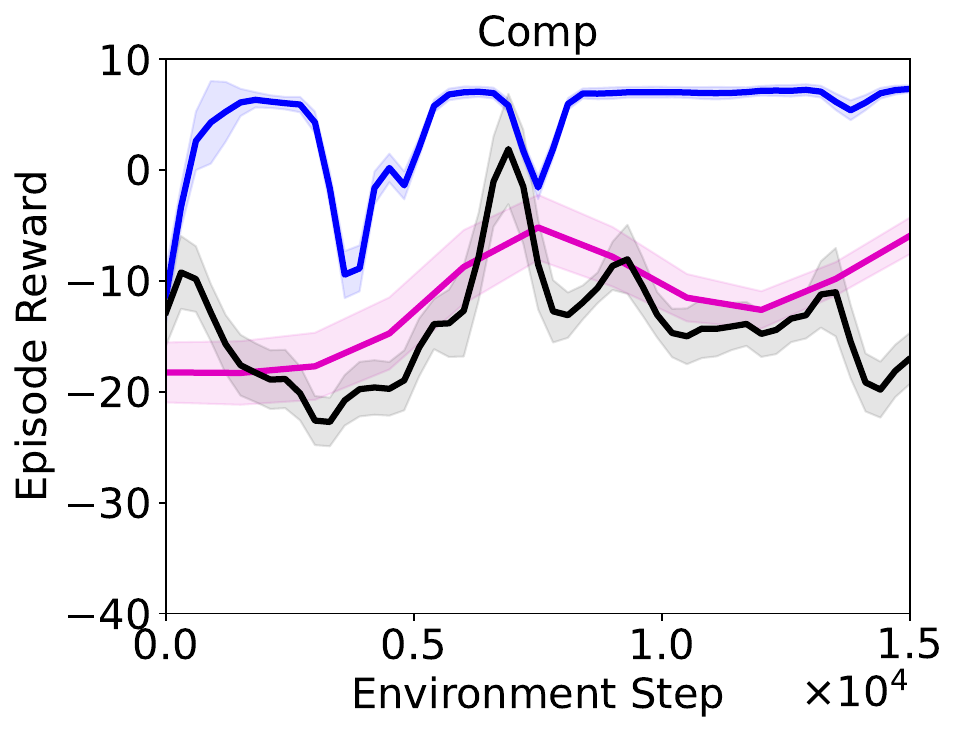}
    \end{tabular}}
    \caption{Learning curves of episode lengths and episode rewards. \coin successfully obtains positive mean episode rewards for all circuits before 12,000 steps (precisely, at 10,500 steps). The solid line
and shaded regions represent the mean and standard deviation, respectively, across ten runs  with random seeds.}
    \label{fig:exp2}

\end{figure*}
\begin{table*}[h!]
\centering
\caption{The First Environment Steps Satsifying Some Criteria Till 200,000 Environment Steps}
\label{tab:my-table}
{
\resizebox{0.95\textwidth}{!}{%
\begin{tabular}{|c|c|c|c|c|c|c|c|}
\hline \diagbox{Criterion}{RL Method} & M3                                     & SAC with Mamba                          & SAC with Transformer & SAC with LSTM & SAC for MT & SAC for ST & PPO for ST \\ \hline
\begin{tabular}[c]{@{}c@{}}Mean Episode  Reward $\geq 0$\\ for All Circuits\end{tabular} & {\color[HTML]{3531FF} \textbf{10,500}} & {\color[HTML]{E000C0} \textbf{150,000}} & Fail                 & Fail          & Fail       & Fail       & Fail       \\ \hline
\begin{tabular}[c]{@{}c@{}}Mean Episode  Length $\leq 25$\\ for All Circuits\end{tabular}  & {\color[HTML]{3531FF} \textbf{10,500}} & {\color[HTML]{E000C0} \textbf{84,000}}  & Fail                 & Fail          & Fail       & Fail       & Fail       \\ \hline
\end{tabular}
}}\vspace{-0.1in}
\end{table*} 
\begin{table}[h!]
\vspace{-0.1in}
\caption{The success rates for each RL algorithm}
\resizebox{\columnwidth}{!}{
\begin{tabular}{|c|c|c|c|}
\hline
                   &                                                                                                       &                                                                                                  &                                                                                       \\
\multirow{-2}{*}{\diagbox{Circuit}{RL Method}} & \multirow{-2}{*}{\begin{tabular}[c]{@{}c@{}}\coin\\ (after 15,000 steps)\end{tabular}} & \multirow{-2}{*}{\begin{tabular}[c]{@{}c@{}}SAC with Mamba\\ (after 150,000 steps)\end{tabular}} & \multirow{-2}{*}{\begin{tabular}[c]{@{}c@{}}MBRL\\ (after 15,000 steps)\end{tabular}} \\ \hline
2SOA               & {\color[HTML]{3531FF} \textbf{7/10}}                                                                  & {\color[HTML]{E000C0} \textbf{5/10}}                                                             & 6/10                                                                                  \\ \hline
R2SOA              & {\color[HTML]{3531FF} \textbf{5/10}}                                                                  & {\color[HTML]{E000C0} \textbf{7/10}}                                                             & 4/10                                                                                  \\ \hline
2STIA              & {\color[HTML]{3531FF} \textbf{10/10}}                                                                 & {\color[HTML]{E000C0} \textbf{10/10}}                                                            & 7/10                                                                                  \\ \hline
Comp               & {\color[HTML]{3531FF} \textbf{10/10}}                                                                 & {\color[HTML]{E000C0} \textbf{10/10}}                                                            & 9/10                                                                                  \\ \hline
\end{tabular}

}\label{tab:sr}\vspace{-0.1in}
\end{table}
\subsection{Experimental Setups}
We prepared four circuits, having different number of target specifications with different circuit parameters, as follows: \textbf{2SOA}: Two-stage operational amplifier; \textbf{C2SOA}: Complementary  of Two-stage operational amplifier by interchanging p-mos and n-mos transistors; \textbf{2STIA}: Two-stage transimpedance amplifier; \textbf{Comp}: Comparator. 

The detailed topologies, circuit parameter ranges, and circuit target specification rages are provided in~TABLE~\ref{tab:my-table} and Fig.~\ref{fig:circuits}. Note that the target specifications vary significantly in both type and number, with 6 for the 2SOA, 2 for the Comp, and 4 for the other circuits. Additionally, for the Comp, the circuit parameters are transistor widths, whereas the other circuits utilize multipliers and capacitance. The total number of circuit parameters is 7 for both the 2SOA and R2SOA, and 6 for the remaining circuits. Consequently, the dimensionality of the observations differs for each circuit, as it is twice the number of target specifications plus the number of circuit parameters.

We choose the following hyperparameters for the environment what we tackle: the episode maximum length $T_{ep}=30$, the number of evaluation $10$, the number of random seeds $10$, and the number of exploration $N_I=3,600$.  TABLE~\ref{tab:hyperparam} shows the hyperparameters of \coin. For experiments, we use NVIDIA GeForce RTX 3060 with AMD Ryzen 9 5950X 16-Core Processor with Ngspice 4.3 with 45nm technology.

\subsection{Main Experiments}\label{sec:exp1}
In our experiments, exploration steps for SAC-related algorithms and corresponding time steps for PPO were excluded to ensure a fair comparison. In addition, as we mentioned, the environment randomly changes circuit types and target specifications. We compare the following baselines to demonstrate the effectiveness of the Mamba architecture: SAC with Mamba, SAC with Transformer, SAC for Multi-Task (MT), SAC for Single-Task (ST), and PPO for Single-Task (ST). SAC for MT refers to training a policy for multi-circuit simultaneously using padded observations, while SAC for ST optimizes each circuit individually. Notice that PPO and SAC for ST only focus on individual circuits, which means that even though the training is successful in the same environment steps under SAC with autoregressive architectures, the sample efficiency is indeed approximately 1/4 lower. SACs with autoregressive architectures focus on multi-circuit optimization. We excluded PPO for MT since it has been reported that SAC for MT outperforms PPO-MT~\cite{yu2020meta}. For the transformer model, we set the model dimension to 128, with 8 attention heads, 1 encoding layer, and a feedforward network dimension of 2,048, utilizing ReLU activations. The layer normalization epsilon is set to \(10^{-5}\). The LSTM networks use input and hidden sizes of 512, with 1 layer. SAC and PPO with fully connected (FC) layers employ hidden layer sizes of (512, 512). Consequently, the number of learnable parameters in the latent representations for SAC with Mamba, SAC with Transformer, SAC with LSTM, SAC with FCs, and PPO with FCs are 437,760, 593,024, 526,336, 525,312, and 525,312, respectively, indicating that SAC with Mamba has the fewest parameters among these models. All other hyperparameters for SAC and PPO not mentioned above follow the defaults in \cite{SpinningUp2018}, produced by OpenAI.

As shown in Fig.~\ref{fig:exp1}, we present the episode reward and length over 10 evaluation runs. SAC and PPO with fully connected layers struggle to outperform SAC with the Mamba architecture, even in single-circuit optimization. SAC with the Transformer underperforms compared to SAC with LSTM, as both fail to efficiently optimize all circuits, unlike SAC with Mamba. We hypothesize that the Transformer's size is insufficient for multi-task learning, and increasing it slows inference, negatively impacting wall-clock performance~\cite{mamba1}. The average inference times for SAC with Mamba and Transformer were $5.101\pm1.499$ms and $44.128\pm4.368$ms on average, respectively, for a batch size of 256 across 100 trials, with the difference growing as network size increases. We speculate that the superior performance arises from Mamba's implicit motivation, which is reflected in the state space model (SSM), a physical system that illustrates state transitions~\cite{mamba1, mamba2}. This SSM closely aligns with the Markov decision process and circuit simulations, suggesting that Mamba may be more suitable than the transformer architecture for analog circuit optimization via reinforcement learning.
In conclusion, Mamba’s fewer learnable parameters and superior sample efficiency make it the preferred choice for further validation.

To show the validness of  MBRL with effective scheduling, we compare \coin with SAC with Mamba and MBRL using the Mamba architecture. For MBRL, we fix the update iteration and the real-synthetic data ratio as~\cite{mbpo}, setting the real-synthetic data ratio 0.05, the number of rollouts 10, and the number of update iterations 20. All other parameters are same with those listed in Table~\ref{tab:my-table}. For \coin, we set the $scale$ 15,000 and moderately increase the number of rollouts from 1 to 7, the real-to-synthetic data ratio in each batch from 0.05 to 0.95, and the number of agent update iterations per environment step from 15 to 20. Using this scheduling, we expect that \coin initially learns to explore using a substantial amount of synthetic data, which may be inaccurate. Over time, it is fine-tuned by more accurate synthetic data alongside a greater amount of real data.

Fig.~\ref{fig:exp2} shows the learning curves of episode lengths and success rates up to 20,000 steps. Although the maximum training timestep is only 20,000, \coin significantly outperforms the other methods until 12,000 environment steps. The order of \( \mathcal{O}(10,000) \) necessary simulator runs is significant due to the computational burden of simulations.

To compare \coin with other RL baselines more rigorously, we provide further experimental results in Table~\ref{tab:my-table} till 200,000 steps. The table shows the first environment steps satisfying specific criteria. Up to 200,000 steps, only Mamba-assisted algorithms (\coin and SAC-Mamba) satisfied the criteria. In particular, \coin outperforms SAC-Mamba by about \(10\times\) in sample efficiency. Finally, TABLE~\ref{tab:sr} shows the success rate after training.  Although \coin is trained only with 15,000 environment steps, it outperforms other algorithms such as SAC with Mamba trained with 150,000 environment steps.

\section{Dicussion}
We proposed a novel model-based RL (MBRL) framework, \coin. This MBRL framework leverages the Mamba architecture, an emerging competitor to the transformer architecture, and MBRL with effective scheduling that gradually adjusts crucial MBRL training  parameters to address the exploration-exploitation dilemma in multi-circuit optimization. To the best of our knowledge, this is the first work to combine these approaches to tackle multi-circuit optimization. \coin outperformed other popular RL baselines such as PPO and SAC in terms of sample efficiency under circuits with distinct circuit parameters and target specifications. Indeed, the policy learned by \coin could optimize all circuits in many cases only with fewer than 15,000 interactions with the real simulator.

\newpage
\bibliographystyle{IEEEtran}

\bibliography{references.bib}

\end{document}